%% file: main.tex
\documentclass[11pt,a4paper]{article}
\usepackage[hyperref]{emnlp2020}

\usepackage{times}
\usepackage{latexsym}

\usepackage{microtype}

\usepackage{setspace}
\usepackage{soul}
\usepackage{url}
\usepackage[greek, english]{babel}
\usepackage[utf8]{inputenc}

\usepackage{color}
\usepackage{comment}
\usepackage{colortbl}

\usepackage{booktabs}
\usepackage{amsfonts}
\usepackage{amsmath}
\usepackage{amssymb}
\usepackage{makecell}
\usepackage{subcaption}
\usepackage{multirow}
\usepackage{pbox}

\usepackage[flushleft]{threeparttable}

\usepackage{graphicx,xspace}
\usepackage{epstopdf}
\usepackage{stfloats}
\usepackage[font=small]{caption}
\usepackage[ruled, vlined, linesnumbered, resetcount, noend]{algorithm2e}
\usepackage{array}
\usepackage{microtype}

\relax

\usepackage{comment}
\usepackage{setspace}

\usepackage{CJKutf8}  
\usepackage{color}
\usepackage{amssymb}
\usepackage{pifont}
\newcommand{\xmark}{\ding{55}}  
\usepackage{svg}

\usepackage{url}
\usepackage{bm}
\usepackage{amsfonts}
\usepackage{amsmath}
\usepackage{amsthm}
\usepackage{stmaryrd}
\usepackage{dsfont}
\usepackage[normalem]{ulem}
\useunder{\uline}{\ul}{}
\usepackage{graphicx}
\usepackage{multirow}
\usepackage{comment}

\usepackage{booktabs}
\usepackage{textcomp}
\usepackage{amssymb}

\newcommand{\tb}{\textbf}
\newcommand{\ud}{\underline}
\definecolor{beige}{rgb}{0.96, 0.96, 0.86}

\newcommand{\tkg}{{i}}  
\newcommand{\skg}{{j}}  

\newcommand{\stitle}[1]{\vspace{0.0ex}\noindent{\bf #1}}
\newcommand{\bh}{\boldsymbol{h}}
\newcommand{\br}{\boldsymbol{r}}
\newcommand{\bt}{\boldsymbol{t}}
\newcommand{\dataone}{DBP-5L}

\newcommand{\alignset}{\Gamma_{G_\tkg \leftrightarrow G_\skg}}

\newcommand{\mj}{\mathcal{J}}
\newcommand{\ml}{\mathcal{J}}
\newcommand{\lpm}{\llbracket p \rrbracket^m}
\newcommand{\modelname}{\texttt{KEnS}}
\newcommand{\boostname}{$\modelname_b$}

\newcommand{\queryexample}{\emph{(The Tale of Genji, genre, \underline{?t})}}

\theoremstyle{definition}
\newtheorem{exmp}{Example}[section]

\title{Multilingual Knowledge Graph Completion via\\ Ensemble Knowledge Transfer}
\author{Xuelu Chen$^1$, Muhao Chen$^{1,2,3}$, Changjun Fan$^4$\thanks{\indent This work was done when this author was visiting University of California, Los Angeles.}\;,\\ 
\textbf{Ankith Uppunda$^1$, Yizhou Sun$^1$ \& Carlo Zaniolo$^1$}\\
$^1$Department of Computer Science, UCLA, USA\\
$^2$Information Sciences Institute, USC, USA\\
$^3$Department of Computer and Information Science, UPenn, USA\\
$^4$College of Systems Engineering, NUDT, China\\
\texttt{shirleychen@ucla.edu}; \texttt{muhaoche@usc.edu};\\ \texttt{fanchangjun@nudt.edu.cn}; \texttt{\{auppunda, yzsun, zaniolo\}@cs.ucla.edu}
}
\date{}

\aclfinalcopy

\begin{document}
\maketitle
\input{0-abstract}
\input{1-introduction}
\input{2-relatedwork}

\input{3-model}

\input{3.3-ensemble}
\input{4-experiments}

\input{5-conclusion}

\bibliographystyle{acl_natbib}
\bibliography{abbr_ref}

\input{6-appendix.tex}
\end{document}

%% file: 0-abstract.tex
\begin{abstract}
Predicting missing facts in a knowledge graph (KG) is a crucial task 
in knowledge base construction and reasoning,
and it has been the subject of much research in
recent works using KG embeddings.
While existing KG embedding approaches mainly learn and predict facts within a single KG,
a more plausible solution would benefit from the knowledge in multiple language-specific KGs,
considering that different KGs have their own strengths and limitations on data quality and coverage.
This is quite challenging, since the transfer of knowledge 
among 
multiple independently maintained KGs is often hindered by the insufficiency of alignment information and the inconsistency of described facts. 
In this paper, we propose \modelname, a novel 
framework for embedding learning and ensemble knowledge transfer 
across
a number of language-specific KGs.
\modelname~embeds all KGs in a shared embedding space, where the association of entities is captured based on self-learning.
Then, \modelname\ 
performs ensemble inference to combine prediction results from embeddings of multiple language-specific KGs,
for which multiple ensemble techniques are investigated.
Experiments on five real-world language-specific KGs 
show that \modelname~consistently improves state-of-the-art
methods on KG completion, via effectively identifying and leveraging complementary knowledge.
\end{abstract}

%% file: 1-introduction.tex
\section{Introduction} \label{sec:introduction}

Knowledge graphs (KGs) store structured representations of real-world entities and relations, constituting actionable knowledge that is crucial to various knowledge-driven applications \cite{koncel2019text,chen2018neural,bordes2014open}. Recently, extensive efforts have been invested in KG embedding models,
which encode entities as low-dimensional vectors and capture relations as algebraic operations on entity vectors.
These models provide a beneficial tool to complete KGs by discovering previously unknown knowledge from latent representations of observed facts.
Representative models including translational models \cite{bordes2013translating,wang2014knowledge} and bilinear models \cite{yang2014embedding,trouillon2016complex} have achieved satisfactory performance in predicting missing facts.

\begin{figure*}[t]
    \centering
    \includegraphics[width=0.99\linewidth]{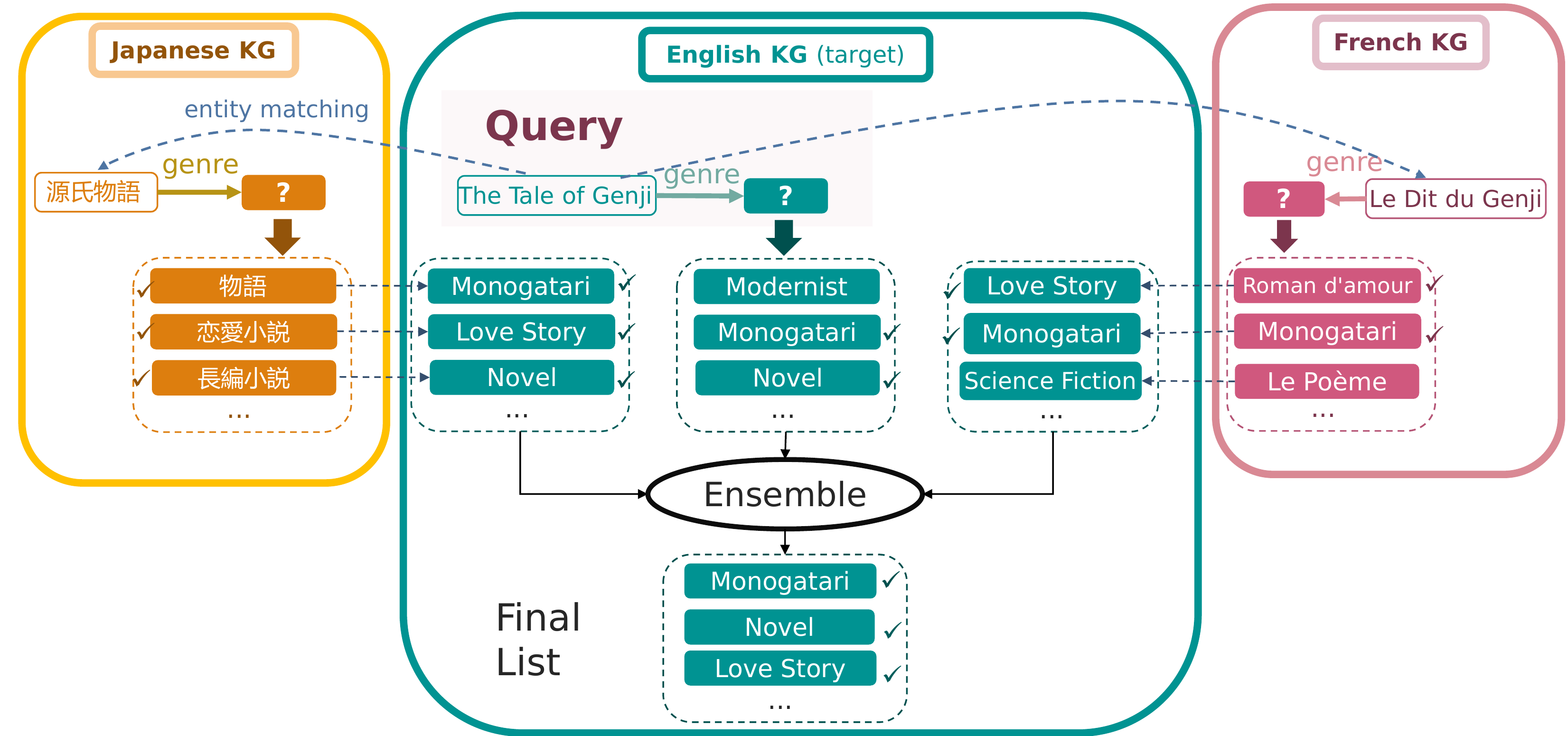}
    \caption{A depiction of the ensemble inference process answering the query \emph{(The Tale of Genji, genre, \underline{?t})} with multiple language-specific KG embeddings. Ground truth answers are marked\emph{Monogatari} is a traditional Japanese literary form.
    } 
    \label{fig:w}\label{fig:overall}
\end{figure*}

Existing methods mainly investigate KG completion within a single monolingual KG.
As different language-specific KGs have their own strengths and limitations on data quality and coverage,
we investigate a more natural solution, which seeks to combine embedding models of multiple KGs in an ensemble-like manner.
This approach offers several potential benefits.
First, embedding models of well-populated KGs (e.g. English KGs) are expected to 
capture richer knowledge 
because of better data quality
and denser graph structures.
Therefore, they would provide ampler signals to facilitate inferring missing facts on sparser KGs.
Second, 
combining the embeddings allows exchanging complementary knowledge across different language-specific KGs. This provides a versatile way of leveraging specific knowledge that is better known in some KGs than the others.
For example, consider the facts about the oldest Japanese novel \textit{The Tale of Genji}.
English DBpedia \cite{lehmann2015dbpedia} only records its genre as \emph{Monogatari} (story), whereas Japanese DBpedia identifies more genres, including \textit{Love Story}, \textit{Royal Family Related Story}, \textit{Monogatari} and \textit{Literature-Novel}.
Similarly, it is reasonable to expect a Japanese KG embedding model to offer significant advantages in inferring knowledge about other Japanese cultural entities such as \textit{Nintendo} and \textit{Mount Fuji}.
Moreover, 
ensemble inference provides a mechanism to assess the credibility of different knowledge sources and thus leads to a more accurate final prediction.

Despite the potential benefits,
combining predictions from multiple KG embeddings represents a non-trivial technical challenge.
On the one hand, knowledge transfer across different embeddings is hindered by the lack of reliable alignment information that bridges different KGs.
Recent works on multilingual KG embeddings provide support for automated entity matching \cite{chen2017multilingual,chen2018co,sun2018bootstrapping,sun2020alinet}. However, the performance of the state-of-the-art (SOTA) entity matching methods is still far from perfect \cite{sun2020alinet}, which may cause erroneous knowledge transfer between two KGs.
On the other hand, independently extracted and maintained language-specific KGs may inconsistently describe some facts,
therefore causing different KG embeddings to give inconsistent predictions and raising a challenge to identifying the trustable sources.
For instance, while the English DBpedia strictly distinguishes the network of a TV series (e.g. BBC) from its channel (e.g. BBC One) with two separate relations, i.e., \texttt{network} and \texttt{channel}, the Greek DBpedia only uses \texttt{channel} to represent all of those.
Another example of inconsistent information is that Chinese DBpedia labels the birth place of the ancient Chinese poet \textit{Li Bai} as \textit{Sichuan, China}, 
which is mistakenly recorded as \textit{Chuy, Kyrgyz} in English DBpedia.
Due to the rather independent extraction process of each KG, such inconsistencies are inevitable, calling upon a reliable approach to identify credible knowledge among various sources.

In this paper, we propose \modelname~(\underline{K}nowledge \underline{Ens}emble),
which, to the best of our knowledge, is the first ensemble framework of KG embedding models.
Fig. \ref{fig:overall} gives a depiction showing the  ensemble inference process of \modelname.
\modelname~seeks to improve KG completion in a multilingual setting, by combining predictions from embedding models of multiple language-specific KGs
and identifying the most probable answers from those prediction results that are not necessarily consistent.
Experiments on five real-world language-specific KGs 
show that \modelname~significantly improves SOTA fact prediction methods that solely rely on a single KG embedding.
We also provide detailed case studies to interpret how a sparse, low-resource KG can benefit from embeddings of other KGs, and how exclusive knowledge in one KG can be broadcasted to others.

%% file: 2-relatedwork.tex
\section{Related Work}

We hereby discuss three lines of work that are closely related to this topic.

\stitle{Monolingual KG Embeddings.}
Monolingual KG embedding models embed entities and relations in a low-dimensional vector space and measure triple plausibility using these vectors.
Translational models assess the plausibility of a triple $(h,r,t)$ by the distance between two entity vectors $\bh$ and $\bt$, after applying a relation-specific translation vector $\br$.
The representative models include TransE \cite{bordes2013translating} and its extensions 
TransD \cite{ji2015knowledge}. 
Despite their simplicity, translational models achieve satisfactory performance on KG completion and are robust against the sparsity of data \cite{hao2019joie}.
RotatE \cite{sun2018rotate} employs a complex embedding space and models the relation $\br$ as the rotation instead of translation of the complex vector $\bh$ toward $\bt$, which leads to the SOTA performance on KG embedding.
There are also various methods falling into the groups of Bilinear models such as RESCAL \cite{nickel2011three} and DistMult \cite{yang2014embedding},
as well as neural models like HolE \cite{nickel2016holographic} and  ConvE \cite{dettmers2017convolutional}.
Due to the large body of work in this line of research, we only provide a highly selective summary here. Interested readers are referred to recent surveys \cite{wang2017knowledge, ji2020survey} for more information.

\stitle{Multilingual KG Embeddings.}
Recent studies have extended embedding models to bridge multiple KGs, typically for KGs of multiple languages. 
MTransE \cite{chen2017multilingual} jointly learns a transformation across two separate translational embedding spaces along with the KG structures.
BootEA \cite{sun2018bootstrapping} introduces a bootstrapping approach to iteratively propose new alignment labels to enhance the performance. 
MuGNN \cite{cao2019multi} encodes KGs via multi-channel Graph Neural Network to reconcile the structural differences.
Some others also leverage side information to enhance the alignment performance, including entity descriptions \cite{chen2018co,zhang2019multi}, attributes \cite{distiawanTrsedya2019,sun2017cross,yang2019aligning}, neighborhood information \cite{wang2018cross,yang2014embedding,li2019semi,sun2019transedge,sun2020alinet} and degree centrality measures \cite{pei2019deg}.
A systematic summary of relevant approaches is given in a recent survey by \citet{sun2020benchmark}.
Although these approaches focus on the KG alignment that is different from the problem we tackle here, such techniques can be leveraged to support entity matching between KGs,
 which is a key component of our framework.

\stitle{Ensemble methods.}
Ensemble learning has been widely used to improve machine learning results by combining 
multiple models on the same task. Representative approaches include voting, bagging \cite{breiman1996bagging}, stacking \cite{wolpert1992stacked} and boosting \cite{freund1997decision}.
Boosting methods seek to combine multiple weak models into a single strong model, particularly by learning model weights from the sample distribution.
Representative methods include AdaBoost \cite{freund1997decision} and RankBoost \cite{freundrankboost2004}, which target at classification and ranking respectively.
AdaBoost starts with a pool of weak classifiers and iteratively selects the best one based on the sample weights in that iteration.
The final classifier is a linear combination of the selected weak classifiers, where each classifier is weighted by its performance. In each iteration, sample weights are updated according to the selected classifier so that the subsequent classifiers will focus more on the \textit{hard} samples.
RankBoost seeks to extend AdaBoost to ranking model combination. The model weights are learned from the ranking performance in a boosting manner.
In this paper, we extend RankBoost to combine ranking results from multiple KG embedding models. This technique addresses KG completion by combining knowledge from multiple sources and effectively compensates for the inherent errors in any entity matching processes.

%% file: 3-model.tex
\section{Method} \label{sec:model}

In this section, we introduce \modelname, 
an embedding-based ensemble inference framework
for multilingual KG completion.

\modelname~conducts two processes, i.e. \emph{embedding learning} and \emph{ensemble inference}.
The embedding learning process trains the \emph{knowledge model} that encodes entities and relations of every KG in a shared embedding space, as well as the \emph{alignment model} that seizes the correspondence in different KGs and enables the projection of queries and answers across different KG embeddings.
The ensemble inference process combines the predictions from multiple KG embeddings to improve fact prediction. 
Particularly, to assess the confidence of predictions from each source, we introduce a boosting method to learn entity-specific weights for knowledge models.

\subsection{Preliminaries}\label{sec:formulation}
A KG $G$ consists of a set of (relational) facts $\{(h,r,t)\}$, where $h$ and $t$ are the head and tail entities of the fact $(h,r,t)$,
and $r$ is a relation. 
Specifically, $h, t \in E$ (the set of entities in $G$), and $r \in R$ (the set of relations). 
To cope with KG completion, the fact prediction task seeks to fill in the right entity for the missing head or tail of an unseen triple.
Without loss of generality, we hereafter discuss the case of predicting missing tails. We refer to a triple with a missing tail as a \emph{query} $q=(h,r,\underline{?t})$.
The answer set $\Omega(q)$ consists of all the right entities that fulfill $q$.
For example, we may have a query \textit{(The Tale of Genji, genre, \underline{?t})}, and its answer set will include \emph{Monogatari}, \emph{Love Story}, etc.

Given KGs in M languages $G_1, G_2, \ldots, G_M$ ($|E_i| \leq |E_j|, i < j$), 
we seek to perform fact prediction on each of those by transferring knowledge from the others.
We consider fact prediction as a ranking task in the KG embedding space, which is to transfer the query to external KGs and to combine predictions from multiple embedding models into a final ranking list.
Particularly, given the existing situation of the major KGs, we use the following settings: 
(i) entity alignment information is available between any two KGs, though limited; and 
(ii) relations in different language-specific KGs 
are represented with a unified schema. 
The reason for the assumption is that unifying relations is usually feasible, since the number of relations is often much smaller compared to the enormous number of entities in KGs. This has been de facto achieved in a number of influential knowledge bases, including DBpedia \cite{lehmann2015dbpedia}, Wikidata \cite{vrandevcic2014wikidata} and YAGO \cite{rebele2016yago}.
In contrast, KGs often consist of numerous entities that cannot be easily aligned, and entity alignment is available only in small amounts.

\subsection{Embedding Learning}
The embedding learning process jointly trains the \emph{knowledge model} and the \emph{alignment model} following \citet{chen2017multilingual}, while self-learning is added to improve the alignment learning.
The details are described below.
 
\stitle{Knowledge model.}
A knowledge model seeks to encode the facts of a KG in the embedding space.  
For each language-specific KG, it characterizes the plausibility of its facts.
Notation-wise, we use boldfaced $\boldsymbol{h}, \boldsymbol{r}, \boldsymbol{t}$ as embedding vectors for head $h$, relation $r$ and tail $t$ respectively.
The learning objective is to minimize the following margin ranking loss:
\begin{equation}\label{eq:jk}
\begin{small}
    \mathcal{J}_K^G = 
    \sum_{(h,r,t)\in G, \atop (h',r,t') \notin G} 
    [f(\bh', \br, \bt') - f(\bh, \br, \bt) + \gamma]_+
\end{small}
\end{equation}
where $[\cdot]_+= \max(\cdot, 0)$, and $f$ is a model-specific triple scoring function. The higher score indicates the higher likelihood that the fact is true. $\gamma$ is a hyperparameter, and $(h', r, t')$ is a negative sampled triple obtained by randomly corrupting either head or tail of a true triple $(h,r,t)$.

We here consider two representative triple scoring techniques:
TransE \cite{bordes2013translating}
and RotatE \cite{sun2018rotate}. 
TransE models relations as translations between head entities and tail entities in a Euclidean space, while RotatE models relations as rotations in a complex space.
The triple scoring functions are defined as follows.
\begin{align}
    f_{\text{TransE}} (h,r,t) &= - \| \bh + \br - \bt \|_2 \\
    f_{\text{RotatE}} (h,r,t) &= - \| \bh \circ \br - \bt \|_2 
\end{align}
where $\circ: \mathbb{C}^d \times \mathbb{C}^d \rightarrow \mathbb{C}^d$ denotes Hadamard product for complex vectors,
and $\|\cdot\|_2$ denotes $L_2$ norm.

\stitle{Alignment model.}
An 
alignment model is trained to match entity counterparts between two KGs on the basis of a small amount of seed entity alignment.
We embed all KGs in one vector space and make each pair of aligned entities embedded closely. 
Given two KGs $G_\tkg$ and $G_\skg$ with $|E_\tkg| \leq |E_\skg|$,
the alignment model loss is defined as:
\begin{equation}
    \mathcal{J}_{A}^{G_\tkg \leftrightarrow G_\skg} = \sum_{(e_\tkg, e_\skg)\in \alignset} \| \boldsymbol{e}_\tkg - \boldsymbol{e}_\skg \|_2
\end{equation}
where $e_\tkg \in E_\tkg, e_\skg \in E_\skg$ and $\alignset$ is the set of seed entity alignment between $G_\skg$ and $G_\tkg$.
Assuming the potential inaccuracy of alignment, we do not directly assign the same vector to aligned entities of different language-specific KGs. 

Particularly, as the seed entity alignment is provided in small amounts,
the alignment process 
conducts self-learning, where training iterations incrementally propose more training data on unaligned entities to guide subsequent iterations.
At each iteration, if a pair of unaligned entities in two KGs are mutual nearest neighbors according to the CSLS measure \cite{conneau2017word}, \modelname~adds this highly confident alignment to the training data. 

\stitle{Learning objective.}
We conduct joint training of knowledge models for multiple KGs and alignment models between each pair of them via minimizing the following loss function:
\begin{equation}\label{eq:loss}
    \ml = \sum_{m=1}^M \mj_{K}^{G_m} 
    + \lambda \sum_{\tkg=1}^M \sum_{\skg=\tkg+1}^M
        \mj_{A}^{G_\tkg \leftrightarrow G_\skg}
\end{equation}

\noindent
where $\mj_{K}^{G_m}$ is the loss of the knowledge model on $G_m$ as defined in Eq (\ref{eq:jk}),
$\mj_{A}^{G_\tkg \leftrightarrow G_\skg}$ is the alignment loss between $G_\tkg$ and $G_\skg$.
$\lambda$ is a positive hyperparameter that weights the two model components.
Following \citet{chen2017multilingual},
instead of directly optimizing $\ml$ in Eq. (\ref{eq:loss}), our implementation optimizes each $\mj_{K}^G$ and each $\lambda \mj_{A}^{G_\tkg \leftrightarrow G_\skg}$ alternately in separate batches.
In addition, we enforce $L_2$-regularization to prevent 
overfitting.

%% file: 3.3-ensemble.tex
\subsection{Ensemble Inference}
We hereby introduce how \modelname\ 
performs fact prediction
on multiple KGs via ensemble inference.

\stitle{Cross-lingual query and knowledge transfer.}
To facilitate the process of completing KG $G_i$ with the knowledge from another KG $G_j$, \modelname~first predicts the alignment for entities between $G_i$ and $G_j$.
Then, it uses the alignment to transfer queries from $G_i$ to $G_j$, and transfer the results back. 
Specifically, alignment prediction is done by performing an kNN search in the embedding space for each entity in the smaller KG (i.e. the one with fewer entities) and find the closest counterpart from the larger KG.
Inevitably, some entities in the larger KG will not be matched with a counterpart due to the 1-to-1 constraint. In this case, we do not transfer queries and answers for that entity.

\stitle{Weighted ensemble inference.}
We denote the embedding models of $G_1, \ldots, G_M$ as $f_1, \ldots, f_M$.
On the target KG where we seek to make predictions, given each query, the entity candidates are ranked by the weighted voting score of the models:
\begin{equation}\label{eq:score}
    s(e) = \sum_{i=1}^M w_i(e) N_i(e)
\end{equation}
where $e$ is an entity on the target KG, and $w_i(e)$ is an entity-specific model weight, $N_i(e)$ is $1$ if $e$ is ranked among top $K$ by $f_i$, otherwise $0$.

We propose three variants of \modelname~that differ in the computing of $w_i(e)$, 
namely $\modelname_b$ , $\modelname_v$ and $\modelname_m$.
Specifically, $\modelname_b$ learns an entity-specific weight $w_i(e)$ for each entity in a \emph{\underline{b}oosting} manner, $\modelname_v$ fixes $w_i(e)=1$ for all $f_i$ and $e$ (i.e. \emph{majority \underline{v}oting)}, and $\modelname_m$ adopts \emph{\underline{m}ean reciprocal rank} (MRR) of $f_i$ on the validation set of the target KG as $w_i(e)$.
We first present the technical details of the boosting-based $\modelname_b$.

\subsubsection{Boosting Based Weight Learning} \label{sec:boost}
$\modelname_b$ seeks to learn model weights for ranking combination,
which aims at reinforcing correct beliefs and compensating for alignment error.
An embedding model that makes more accurate predictions should receive a higher weight.
Inspired by RankBoost \cite{freundrankboost2004}, we reduce the ranking combination problem to a classifier ensemble problem.
$\modelname_b$ therefore learns model weights in a similar manner as AdaBoost.

\stitle{Validation queries and critical entity pairs.}
To compute entity-specific weights $w_i(e)$, \boostname~evaluates the performance of $f_i$ on a set of \textit{validation queries} related to $e$. These queries are converted from all the triples in the validation set that mention $e$.
An example of validation queries for the entity \emph{The Tale of Genji} is given as below.

\begin{exmp} Examples of triples and validation queries for the entity \emph{The Tale of Genji}.
\begin{small}
\begin{align*}
&\textbf{Triples:}  \\
& \{\texttt{(The Tale of Genji, country, Japan)}\\
    & \texttt{(The Tale of Genji, genre, Monogatari)} \\
    & \texttt{(The Tale of Genji, genre, Love Story)} \}\\
& \textbf{Queries:} \\
& Q=\{q_1=\texttt{(The Tale of Genji, country, \underline{?t})}\\
    & \quad \quad q_2=\texttt{(The Tale of Genji, genre, \underline{?t})} \}
\end{align*}
\end{small}
\end{exmp}

\noindent
Similar to RankBoost \cite{freundrankboost2004}, given a query $q$, $\modelname_b$~evaluates the ranking performance of a model by checking if each of the \emph{critical entity pairs} \{$(e,e')$\} is ranked in correct order, where $e$ is a ground truth tail and $e'$ is an incorrect one. 
An example of critical entity pairs is given as below:
\begin{exmp} Critical entity pairs for the query \queryexample. Ground truth tails are boldfaced. Pairs with x-marks indicate wrong prediction orders.
\begin{small}
\begin{align*}
    & \text{\underline{Correct ranking}}: \\ 
    & \text{\textbf{Monogatari, Love Story}, Modernist, Science Fiction} \\
    & \text{\underline{Predicted ranking}:}\\ 
    & \text{Modernist, \textbf{Monogatari, Love Story}, Science Fiction}\\
    & \text{\underline{Critical pair ranking results:}} \\
    & \text{(\textbf{Monogatari}, Modernist) \xmark}, \text{(\textbf{Love Story}, Modernist) \xmark} \\
    & \text{(\textbf{Monogatari}, Science Fiction) \checkmark},\\ 
    & \text{(\textbf{Love Story}, Science Fiction) \checkmark} \\
    & \text{\underline{Uncritical pairs:}}\\
    & \text{(Monogatari, Love Story), (Modernist, Science Fiction)}
\end{align*}
\end{small}
\end{exmp}


\stitle{Ranking loss.}
The overall objective of \boostname~is to minimize the sum of ranks of all correct answers in the combined ranking list
$\sum_{q} \sum_{e \in \Omega(q)}r(e)$,
where $\Omega(q)$ is the answer set of query $q$ and $r(e)$ is the rank of entity $e$ in the combined ranking list of the ensemble inference.
Essentially, the above objective is minimizing the number of mis-ordered critical entity pairs in the combined ranking list.
Let the set of all the critical entity pairs from all the validation queries of an entity as $P$.
\citet{freundrankboost2004} have proved that, when using RankBoost, this ranking loss is bounded as follows:
\begin{equation*}
    | \{ p: p \in P, p~\text{is mis-ordered} \} |  
    \leq |P| \prod_{m=1}^M Z^m
\end{equation*}
where 
$M$ is the number of KGs and therefore the maximum number of rounds in boosting. $Z^m$ is the weighted ranking loss of the $m$-th round:
\begin{equation} \label{eq:zm}
    Z^m =  \sum_{p \in P}
    D^m(p) e^{- w^m \lpm}
\end{equation}
where $\lpm = 1$ if the critical entity pair $p$ is ranked in correct order by the selected embedding model in the $m$-th round, otherwise $\lpm = -1$, 
$D^m(p)$ is the weight of the critical entity pair $p$ in the $m$-th round, and $w^m$ is 
the weight of the chosen model in that round.
Now the ranking combination problem is reduced to a common classifier ensemble problem. 

\stitle{Boosting procedure.}
The boosting process alternately repeats two steps:
(i) Evaluate the ranking performance of the embedding models and choose the best one $f^m$
according to the entity pair weight distribution in that round;
(ii) Update entity pair weights to put more emphasis on the pairs which $f_m$ ranks incorrectly.

Entity pair weights are initialized uniformly over $P$ as $D^1(p) = \frac{1}{|P|}, p \in P$.
In the $m$-th round ($m=1,2,...,M)$, \boostname~chooses an embedding model $f^m$ and sets its weight $w^m$, seeking to minimize the weighted ranking loss $Z^{m}$ defined in Eq.(\ref{eq:zm}). By simple calculus, when choosing the embedding model $f_i$ as the model of the $m$-th round, $w^m_i$ should be set as follows to minimize $Z^m$:
\begin{equation} \label{eq:w}
    w^m_i = \frac{1}{2} \ln ( 
    \frac{\sum_{p \in P, \llbracket p \rrbracket = 1}D^m(p)}
    {\sum_{p \in P, \llbracket p \rrbracket = -1}D^m(p)}
    )
\end{equation}
As we can see from Eq. (\ref{eq:w}), the higher $w^m_i$ indicates the better performance of $f_i$ under the current entity pair weight distribution $D^m$.
We select the best embedding model in the $m$-th round $f^m$ based on the maximum weight $w^m=\max\{w^m_1, ..., w^m_M\}$. 

After choosing the best model $f^m$ at this iteration, we update the entity pair weight distribution to put more emphasis on what $f^m$ ranked wrong.
The new weight distribution $D^{m+1}$ is updated as:
\begin{equation}
    D^{m+1}(p) =
    \frac{1}{Z^m} D^m(p) e^{-w^m \lpm}
\end{equation}
where $Z^m$ works as a normalization factor. \boostname~decreases the weight of $D(p)$ if the selected model ranks the entity pair in correct order and increases the weight otherwise. Thus, $D(p)$ will tend to concentrate on the pairs whose relative ranking is hardest to determine.

For queries related to a specific entity, this process is able to recognize the embedding models that perform well on answering those queries and rectify the mistakes made in the previous iteration.

\subsubsection{Other Ensemble Techniques} \label{sec:votemrr}

We also 
investigate two other model variants with
simpler ensemble techniques.

\stitle{Majority vote} ($\modelname_v$):
A straightforward ensemble method is 
to re-rank entities by their nomination counts in the prediction of all knowledge models, which substitutes the voting score (Eq.~\ref{eq:score}) with $s(e) = \sum_{i=1}^M {N_i(e)}$,
\noindent
where $N_i(e)$ is 1 if $e$ is ranked among the top $K$ by the knowledge model $f_i$, otherwise 0. When there is a tie, we order by the MRR given by the models on the validation set.

\stitle{MRR weighting} ($\modelname_m$):
MRR is a widely-used metric for evaluating the ranking performance of a model \cite{bordes2013translating,yang2014embedding,trouillon2016complex},
which may also serve as a weight metric for estimating the prediction confidence of each language-specific embedding in ensemble inference \cite{shen2017setexpan}. 
Let the MRR of $f_i$ be $u_i$ on the validation set, the entities are ranked according to the weighted voting score $s(e) = \sum_{i=1}^M  u_i {N_i(e)}$. 

%% file: 4-experiments.tex
\section{Experiments}

\begin{table}[t]
\centering
\footnotesize
\setlength{\tabcolsep}{5pt}
\begin{tabular}{c|ccccc}
\hline
\hline
Lang. & EN & FR & ES & JA & EL\\
\hline
\#Ent. & 13,996 & 13,176 & 12,382 & 11,805 & 5,231 \\
\#Rel. & 831 & 178 & 144 & 128 & 111 \\
\#Triples & 80,167 & 49,015 & 54,066 & 28,774 & 13,839\\
\hline
\hline
\end{tabular}
\caption{Statistics of \dataone dataset. \textit{Ent.} and \textit{Rel.} stand for entities and relations respectively.}
\label{table:data1}
\end{table}


In this section, we conduct the experiment of fact prediction by comparing \modelname\ variants with various KG embeddings.
We also provide a detailed case study to help understand the principle of ensemble knowledge transfer.



\input{table/4.2resulttable.tex}

\subsection{Experiment Settings}
To the best of our knowledge, existing datasets for fact prediction contain only one monolingual KG or bilingual KGs. Hence, we 
prepared
a new dataset \dataone, which contains five language-specific KGs extracted from English (EN), French (FR), Spanish (ES) and Japanese (JA) and Greek (EL) DBpedia \cite{lehmann2015dbpedia}.
Table~\ref{table:data1} lists the statistics of the contributed dataset DBP-5L.
The relations of the five KGs are represented in a unified schema, which is consistent with the problem definition in Section~\ref{sec:formulation}. The English KG is the most populated one among the five.
To produce KGs with a relatively consistent set of entities,
we induce the subgraphs by starting from a set of seed entities 
where we have 
alignment among 
 all language-specific KGs and
then incrementally collecting triples that involve other entities.
Eventually between any two KGs, the alignment information covers around 40\% of entities.
Based on the same set of seed entities, the Greek KG ends up with 
a notably smaller vocabulary
and fewer triples than the other four. 
We split the facts in each KG into three parts: 60\% for training, 30\% for validation and weight learning, and 10\% for testing.

\stitle{Experimental setup.}
We use the Adam \cite{kingma2014adam} as the optimizer and fine-tune the hyper-parameters by grid search based on $Hits@1$ on the validation set. 
We select among the following sets of hyper-parameter values:
learning rate $lr \in \{0.01, 0.001, 0.0001 \}$, dimension $d \in \{64, 128, 200, 300 \}$, batch size $b \in \{256, 512, 1024 \}$, and TransE margin $\gamma \in \{0.3, 0.5, 0.8\}$.
The best setting is \{$lr=0.001$, $d=300$, $b=256$\} for $\modelname$(TransE) and \{$lr=0.01$, $d=200$, $b=512$\} for $\modelname$(RotatE).
The margin for TransE is $0.3$. The $L_2$ regularization coefficient is fixed as $0.0001$.

\stitle{Evaluation protocol.}
For each test case $(h,r,t)$, we consider it as a query $(h,r,\underline{?t})$ and retrieve top $K$ prediction results for $\underline{?t}$. We compare the proportion of queries with correct answers ranked within top $K$ retrieved entities. We report three metrics with $K$ as $1,3,10$.
$Hits@1$ is equivalent to accuracy.
All three metrics are preferred to be higher.
Although another common metric, Mean Reciprocal Rank (MRR), has been used in previous works \cite{bordes2013translating}, it is not applicable to the evaluation of our framework because our ensemble framework combines the top entity candidates from multiple knowledge models and yields top $K$ final results without making any claims for entities out of this scope.
Following previous works, we use the ``filtered'' setting with the premise that the candidate space has excluded the triples that have been seen in the training set \cite{wang2014knowledge}.

\stitle{Competitive methods.}
We compare six variants of \modelname, which are 
generated by combining
two knowledge models and three ensemble inference techniques introduced in in Section~\ref{sec:model}.
For baseline methods, besides the single-embedding TransE \cite{bordes2013translating}
and RotatE \cite{sun2018rotate}, we also include DistMult \cite{yang2014embedding}, TransD \cite{ji2015knowledge},
 and HolE \cite{nickel2016holographic}.
 After extensive hyperparameter tuning, the baselines are set to their best configurations.
 We also include a baseline named \emph{RotatE+PARIS}, which trains RotatE on 5 KGs and uses the representative non-embedding symbolic entity alignment tool PARIS \cite{suchanek2011paris} for entity matching. PARIS delivered entity matching predictions for 58\%-62\% entities in the English, French, and Spanish KG,
 but almost no matches are delivered for entities in the Greek and Japanese KG, since PARIS mainly relies on entity label similarity. The results on the Greek and Japanese KG are thus omitted for RotatE+PARIS.

\input{4.2-mainresults}

\input{4.3-casestudy}

%% file: table/4.2resulttable.tex
\setlength\tabcolsep{2pt}
\begin{table*}[t]
\centering
\begin{tabular}{c|ccc|ccc|ccc|ccc|ccc}
\hline
\hline
\multicolumn{1}{c|}{KG}            
& \multicolumn{3}{c|}{Greek}                                                
& \multicolumn{3}{c|}{Japanese}                                              
& \multicolumn{3}{c|}{Spanish}                                              
& \multicolumn{3}{c|}{French}                                             
& \multicolumn{3}{c}{English}                                                    \\ \hline
Hits@k (\%)             
& 1                  & 3                   & 10 
& 1                  & 3                   & 10   
& 1                  & 3                   & 10  
& 1                  & 3                   & 10
& 1                  & 3             & 10                   \\
\hline \hline

TransD                  
 & 2.8     & 16.9      & 29.8  
 & 4.2     & 16.3      & 28.8             
& 2.12   & 20.4        & 11.5                      
 & 3.3     & 14.4     & 25.7                      
& 2.9     & 15.4       & 27.4                   
\\
DistMult                
 & 8.9           & 13.0       & 11.3
& 9.3            & 18.4       & 27.5            
   & 7.4         & 15.0       & 22.4                  
 & 6.1                & 14.3          & 23.8 
 & 8.8                & 19.4          & 30.0       
\\
HolE
& 4.2 & 9.5 & 18.3 
& 25.5 &  29.5 & 32.8 
& 20.1 & 26.8 & 29.4 
& 22.4 & 24.4 & 28.9 
& 12.3 & 20.4 & 25.4 \\
\hline
TransE                  
& 13.1   & 23.4   & 43.7 
& 21.1   & 34.4  & 48.5
& 13.5  & 29.4  & 45.0
& 17.5   & 33.1  & 48.8 
& 7.3  & 16.4    & 29.3 
 \\
 \rowcolor[gray]{.9}
$\modelname_v$(TransE)  
&  23.1   &   36.7      &   64.7     
&  22.6       &   35.2     &   52.5             
 &   15.0        &    28.3    &   \textbf{49.0}        
&  18.7       &   29.4       &  52.0    
&   10.8       &  20.4       &  \textbf{39.4}                
\\   
\rowcolor[gray]{.9}
$\modelname_m$(TransE)         
&    26.3     &    42.1      &    65.8  
&   26.1      &   37.7      &    55.3        
 &    16.8    &  \tb{32.9}        &  48.6           
 &  20.5      &   35.6     &  52.8     
&  11.4       & 21.2       & 31.3                
\\
\rowcolor[gray]{.9}
$\modelname_b$(TransE)       
&   \tb{26.4}     &    \ud{\tb{42.4}}      &   \ud{\tb{66.1}} 
&   \tb{26.7}        &   \tb{39.8}      &   \tb{56.4}          
 & \tb{17.4}         &   \tb{32.6}      &    48.3        
 &  \tb{20.8}     &     \tb{35.9}     &   \tb{53.1}
&    \tb{11.7}        &     \tb{21.8}    &  32.0                
\\


\hline
RotatE         
&  14.5   &    18.8    &   36.2     
 &     26.4       &   36.2      &   60.2  
&  21.2     &   31.6      &    53.9           
&  23.2       &    29.4    &    55.5    
&   12.3     &    25.4    &   30.4
\\
RotatE+PARIS         
&  - & - & -   
 &    - & - & -  
&  20.8    &   39.4     &    59.1           
&  22.8       &    32.4    &    60.8   
&   12.4     &    22.7    &   31.5
\\
\rowcolor[gray]{.9}
$\modelname_v$(RotatE)
&    20.5   &   34.3       &  50.1   
&  31.9       &     50.0     &    65.0    
&    20.8       &  41.0         &    59.9 
 &     23.7      &   42.7      &    61.9          
&   13.4     & 23.6    &    34.2 
\\
\rowcolor[gray]{.9}
$\modelname_m$(RotatE)
&    22.0    &     35.0    &  51.4   
& 32.0     &     49.9      &    65.0    
&  21.2    &   41.6    &    60.0  
&  24.5      &   44.8      &   62.5   
 &   12.1          &   24.5  &   34.3           
 \\
\rowcolor[gray]{.9}
$\modelname_b$(RotatE)        
&  \ud{\tb{27.5}}   &    \tb{40.6}      &  \tb{56.5} 
&  \ud{\tb{32.9}}          &  \ud{\tb{49.9}}       &  \ud{\tb{64.8}}         
&    \ud{\tb{22.3}}         &    \ud{\tb{42.4}}     &  \ud{\tb{60.6}}  
&    \ud{\tb{25.2}}        &   \ud{\tb{  44.5 }}   &   \ud{\tb{62.6  }}    
&    \ud{\tb{  14.4}}   &   \ud{\tb{   27.0  }}     &   \ud{\tb{ 39.6   }}              \\
\hline\hline
\end{tabular}

\caption{Fact prediction results on DBP-5L. The overall best results are under-scored.
}
\label{tb:results}

\end{table*}

%% file: 4.2-mainresults.tex
\subsection{Main Results}
The results are reported in Table \ref{tb:results}.
As shown, the ensemble methods by \modelname~lead to consistent improvement in fact prediction. 
Overall, the ensemble inference leads to 1.1\%-13.0\% of improvement in $Hits@1$ over the best baseline methods.
The improved accuracy shows that it is effective to leverage complementary knowledge from external KGs for KG completion.
We also observe that \modelname~brings larger gains on sparser KGs than on the well-populated ones. 
Particularly, on the low-resource Greek KG, $\modelname_b$(RotatE) improves $Hits@1$ by as much as 13.0\% over its single-KG counterpart. 
This finding corroborates our intuition that the KG with lower knowledge coverage and sparser graph structure benefits more from complementary knowledge.

\begin{figure*}[t]
    \centering
    \hspace{6.2em}
    \includegraphics[width=0.61\linewidth]{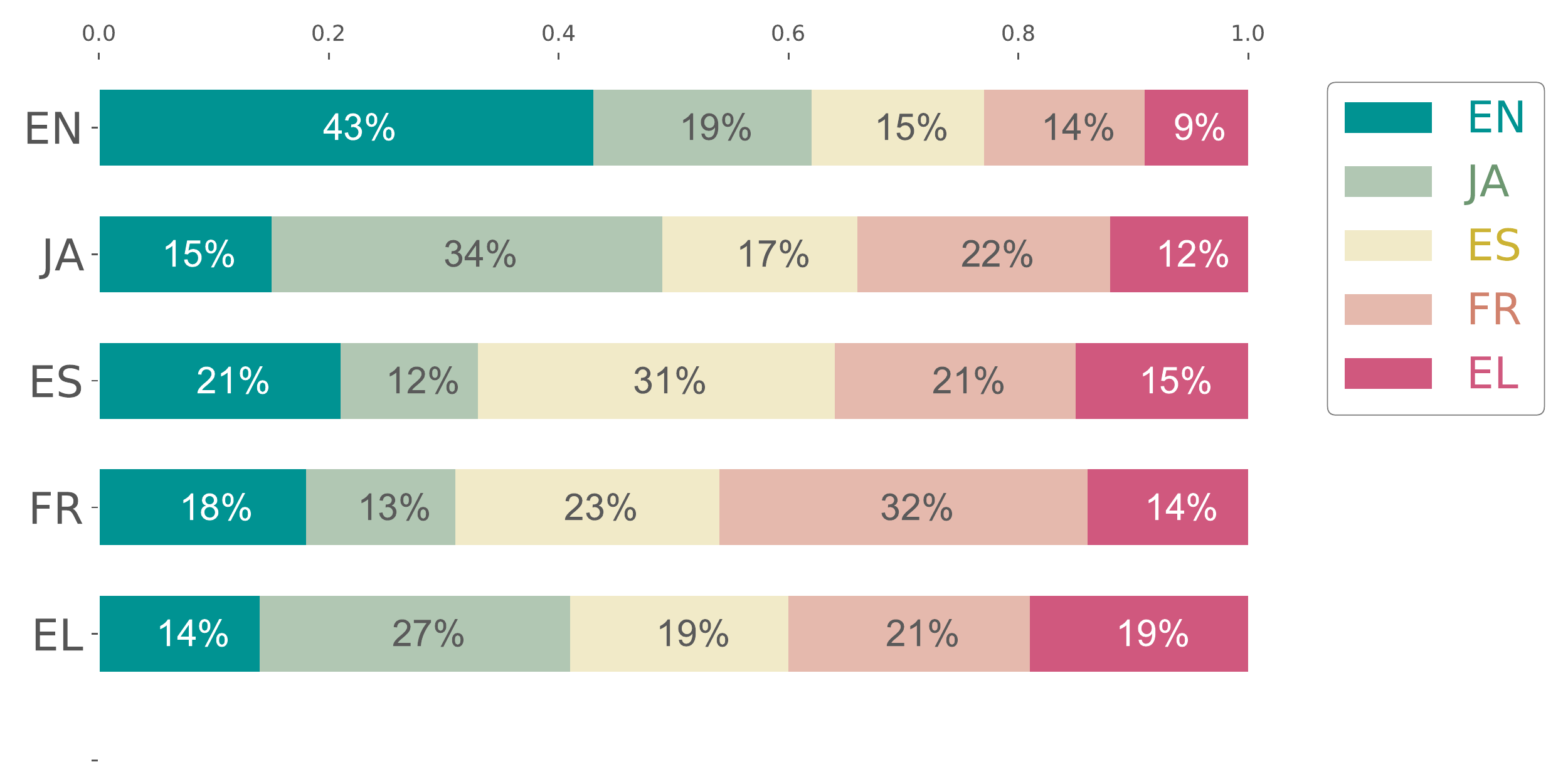}
    \vspace{-2em}
    \caption{Average model weights learned by $\modelname_b$(TransE).} 
    \label{fig:avgw}
\end{figure*}

\begin{figure*}[t]
    \centering
    \includegraphics[width=0.75\linewidth]{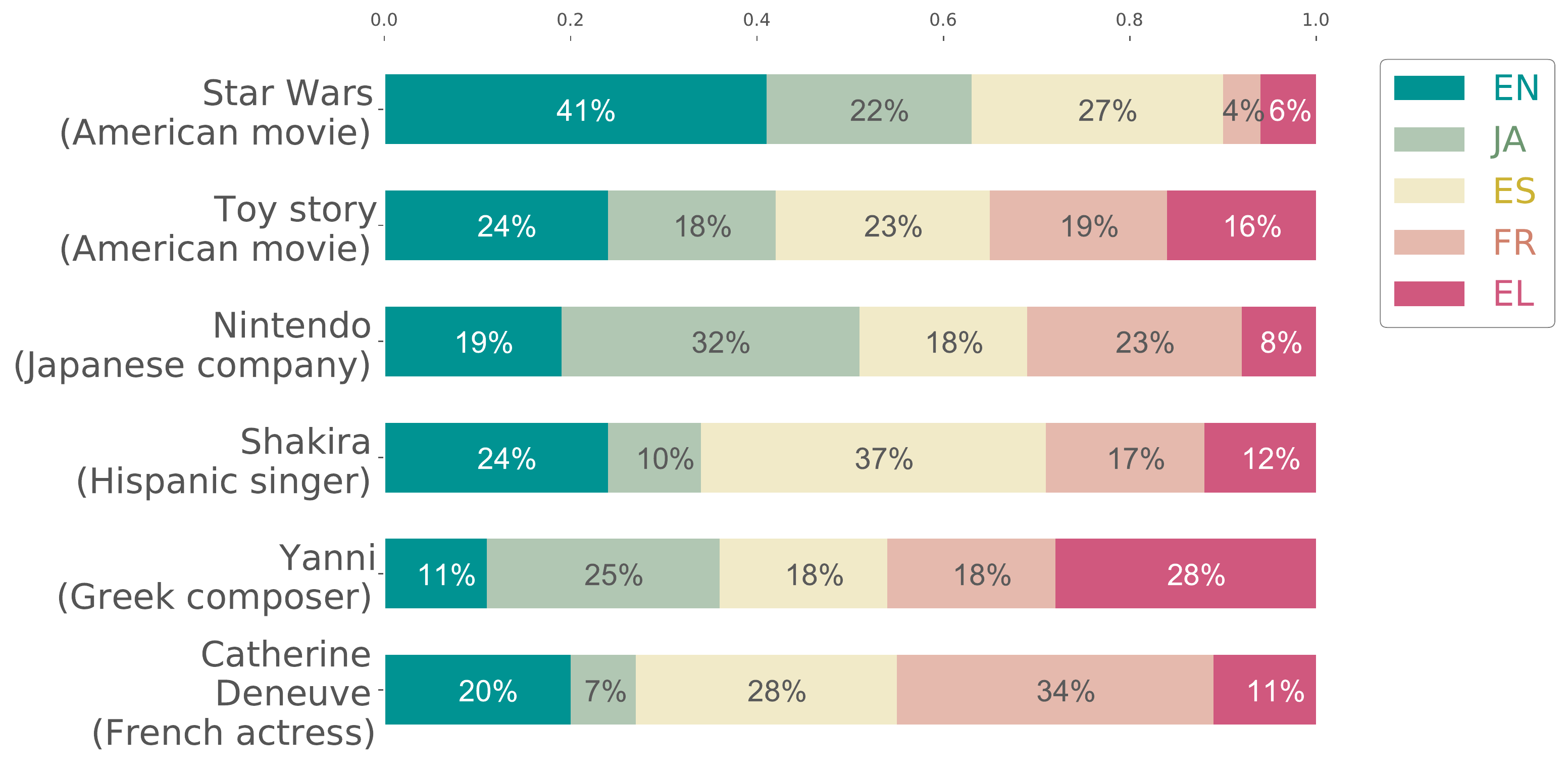}
    \caption{Examples of language-specific model weights learned by $\modelname_b$(TransE). Percentages have been rounded.
    } 
    \label{fig:w}
\end{figure*}

Among the variants of ensemble methods, $\modelname_m$ offers better performance than $\modelname_v$, and $\modelname_b$ outperforms the other two in general.
For example, on the Japanese KG, $\modelname_v$(TransE) improves $Hits@1$ by 3.5\% from the single-KG TransE, while $\modelname_m$ leads to a 5.0\% increase, and $\modelname_b$ further provides a 5.6\% of improvement.
The results suggest that MRR is an effective measure of the trustworthiness of knowledge models during ensemble inference. 
Besides, $\modelname_b$ is able to assess  trustworthiness at a finer level of granularity by learning entity-specific model weights and can thus further improve the performance.

In summary, the promising results by \modelname\ variants show the effectiveness of transferring and leveraging cross-lingual knowledge for KG completion.
Among the ensemble techniques, the boosting technique represents the most suitable one for combining the prediction results from different models.




%% file: 4.3-casestudy.tex
\subsection{Case Studies}
In this section, we provide case studies 
to show how
$\modelname$ is able to transfer cross-lingual knowledge to populate different KGs.

\stitle{Model weights.}
The key to the significantly enhanced performance of $\modelname_b$ is the effective combination of multilingual knowledge from multiple sources. Fig \ref{fig:avgw} shows the average model weight learnt by $\modelname_b$(TransE), which depicts how external knowledge from cross-lingual KGs contributes to target KG completion in general.
The model weights imply that sparser KGs benefit more from the knowledge transferred from others. 
Particularly, when predicting for the Greek KG, the weights of other languages sums up to 81\%. This observation indicates that the significant boost received on the Greek KG comes with the fact that it has accepted the most complementary knowledge from others.
In contrast, when predicting on the most populated English KG, the other language-specific models give a lesser total weight of 57\%.

\input{table/4.3casestudytable}

Among the three KEns variants, the superiority of $\modelname_b$ is attributed to identification of more credible knowledge sources, thus making more accurate predictions. For language-specific KGs, the higher level of credibility often stems from the cultural advantage the KG has over the entity.
Fig \ref{fig:w} presents the model weights for 6 culture-related entities learned by $\modelname_b$(TransE). It shows that KEns can locate the language-specific knowledge model that has a cultural advantage and assign it with a higher weight, which is the basis of an accurate ensemble prediction.

\stitle{Ensemble inference.}
To help understand how the combination of multiple KGs improves KG completion and show the 
effectiveness of leveraging complementary culture-specific knowledge ,
we present a case study about predicting the fact \texttt{(Nintendo, industry, \ud{?t})} for English KG.
Table \ref{tab:tails} lists the top 3 predicted tails yielded by the $\modelname$(TransE) variants,
along with those by the English knowledge model and supporter knowledge models before ensemble.
The predictions made by the Japanese KG are the closest to the ground truths. The reason may be that Japanese KG has documented much richer knowledge about this Japanese video game company, including many of the video games that this company has released.
Among the three \modelname~variants, $\modelname_b$ correctly identifies Japanese as the most credible source and yields the best ranking.

%% file: table/4.3casestudytable.tex
\begin{table}[t]
\footnotesize
\begin{tabular}{c|l}
\hline \hline
Model      & Top 3 Predicted Tails                                                \\
\hline
English 
& Television, Publishing, Information technology
\\
Japanese 
& \textbf{Video game}, Anime, \textbf{Consumer electronics}
\\
Spanish
& Music, Telecommunication, Retail
\\
French 
& Retail,  Television, \textbf{Video game},
\\
Greek
& Nintendo, Music, Wii
\\
\hline
$\modelname_v$
& [\textbf{Video game}, Television](tie), Music
\\
\hline
$\modelname_m$
& Television, \textbf{Video game}, Music
\\
\hline
$\modelname_b$
& \textbf{Video game}, Television, \textbf{Consumer electronics}
\\
\hline \hline
\end{tabular}
    \caption{An example of fact prediction on the English KG by the English knowledge model, four supporter knowledge models, and \modelname(TransE)~variants. Top 3 predicted tails for the query \texttt{(Nintendo, industry, \underline{?t})} are listed. Ground truths are boldfaced.
    }
    \label{tab:tails}
\end{table}

%% file: 5-conclusion.tex
\section{Conclusion}
In this paper, 
we have proposed a new ensemble prediction framework
aiming at collaboratively predicting unseen facts using embeddings of different language-specific KGs.
In the embedding space, our approach jointly captures both the structured knowledge of each KG and the entity alignment that bridges the KGs. 
The significant performance improvements delivered by our model on the task of KG completion were demonstrated by extensive experiments.
This work also suggests promising directions of
future research. One 
is to exploit the potential of \modelname\ on completing low-resource KGs, and
the other is to extend the ensemble transfer mechanism to population sparse domain knowledge in biological \cite{hao2020biojoie} and medical knowledge bases \cite{zhang2020diagnostic}.
Pariticularly, we also seek to ensure the global logical consistency of predicted facts in the ensemble process by incorporating probabilistic constraints \cite{chen2019uncertain}.



\section*{Acknowledgement}

We appreciate the anonymous reviewers for their insightful comments.
Also, we would like to thank Junheng Hao for helping with proofreading the manuscript.

This research is supported in part by Air Force Research Laboratory under agreement number FA8750-20-2-10002. The U.S. Government is authorized to reproduce and distribute reprints for Governmental purposes notwithstanding any copyright notation thereon. The views and conclusions contained herein are those of the authors and should not be interpreted as necessarily representing the official policies or endorsements, either expressed or implied, of Air Force Research Laboratory or the U.S. Government.

%% file: main.bbl
\begin{thebibliography}{43}
\expandafter\ifx\csname natexlab\endcsname\relax\def\natexlab#1{#1}\fi

\bibitem[{Bordes et~al.(2013)Bordes, Usunier, Garcia-Duran, Weston, and
  Yakhnenko}]{bordes2013translating}
Antoine Bordes, Nicolas Usunier, Alberto Garcia-Duran, Jason Weston, and Oksana
  Yakhnenko. 2013.
\newblock Translating embeddings for modeling multi-relational data.
\newblock In \emph{Advances in Neural Information Processing Systems (NIPS)},
  pages 2787--2795.

\bibitem[{Bordes et~al.(2014)Bordes, Weston, and Usunier}]{bordes2014open}
Antoine Bordes, Jason Weston, and Nicolas Usunier. 2014.
\newblock Open question answering with weakly supervised embedding models.
\newblock In \emph{Proceedings of the European Conference on Machine Learning
  and Knowledge Discovery in Databases (ECML-PKDD}, volume 8724, pages
  165--180. Springer.

\bibitem[{Breiman(1996)}]{breiman1996bagging}
Leo Breiman. 1996.
\newblock Bagging predictors.
\newblock \emph{Machine learning}, 24(2):123--140.

\bibitem[{Cao et~al.(2019)Cao, Liu, Li, Li, and Chua}]{cao2019multi}
Yixin Cao, Zhiyuan Liu, Chengjiang Li, Juanzi Li, and Tat-Seng Chua. 2019.
\newblock Multi-channel graph neural network for entity alignment.
\newblock In \emph{Proceedings of the Annual Meeting of Associations for
  Computational Linguistics (ACL)}, pages 1452--1461. Association for
  Computational Linguistics.

\bibitem[{Chen et~al.(2018{\natexlab{a}})Chen, Meng, Huang, and
  Zaniolo}]{chen2018neural}
Muhao Chen, Changping Meng, Gang Huang, and Carlo Zaniolo. 2018{\natexlab{a}}.
\newblock Neural article pair modeling for wikipedia sub-article matching.
\newblock In \emph{Joint European Conference on Machine Learning and Knowledge
  Discovery in Databases}, pages 3--19. Springer.

\bibitem[{Chen et~al.(2018{\natexlab{b}})Chen, Tian, Chang, Skiena, and
  Zaniolo}]{chen2018co}
Muhao Chen, Yingtao Tian, Kai-Wei Chang, Steven Skiena, and Carlo Zaniolo.
  2018{\natexlab{b}}.
\newblock Co-training embeddings of knowledge graphs and entity descriptions
  for cross-lingual entity alignment.
\newblock In \emph{Proceedings of the International Joint Conference on
  Artificial Intelligence (IJCAI)}, pages 3998--4004. International Joint
  Conferences on Artificial Intelligence Organization.

\bibitem[{Chen et~al.(2017)Chen, Tian, Yang, and
  Zaniolo}]{chen2017multilingual}
Muhao Chen, Yingtao Tian, Mohan Yang, and Carlo Zaniolo. 2017.
\newblock Multilingual knowledge graph embeddings for cross-lingual knowledge
  alignment.
\newblock In \emph{Proceedings of the International Joint Conference on
  Artificial Intelligence (IJCAI)}, pages 1511--1517. International Joint
  Conferences on Artificial Intelligence.

\bibitem[{Chen et~al.(2019)Chen, Chen, Shi, Sun, and
  Zaniolo}]{chen2019uncertain}
Xuelu Chen, Muhao Chen, Weijia Shi, Yizhou Sun, and Carlo Zaniolo. 2019.
\newblock Embedding uncertain knowledge graphs.
\newblock In \emph{Proceedings of AAAI Conference on Artificial Intelligence
  (AAAI)}, pages 3363--3370. {AAAI} Press.

\bibitem[{Conneau et~al.(2018)Conneau, Lample, Ranzato, Denoyer, and
  J{\'e}gou}]{conneau2017word}
Alexis Conneau, Guillaume Lample, Marc'Aurelio Ranzato, Ludovic Denoyer, and
  Herv{\'e} J{\'e}gou. 2018.
\newblock Word translation without parallel data.
\newblock In \emph{International Conference on Learning Representations
  (ICLR)}.

\bibitem[{Dettmers et~al.(2018)Dettmers, Minervini, Stenetorp, and
  Riedel}]{dettmers2017convolutional}
Tim Dettmers, Pasquale Minervini, Pontus Stenetorp, and Sebastian Riedel. 2018.
\newblock Convolutional 2d knowledge graph embeddings.
\newblock In \emph{Proceedings of AAAI Conference on Artificial Intelligence
  (AAAI)}, pages 1811--1818. {AAAI} Press.

\bibitem[{Freund et~al.(2004)Freund, Iyer, Schapire, and
  Singer}]{freundrankboost2004}
Yoav Freund, Raj Iyer, Robert~E. Schapire, and Yoram Singer. 2004.
\newblock {RankBoost: An efficient boosting algorithm for combining
  preferences}.
\newblock \emph{Journal of Machine Learning Research (JMLR)}, 4(6):933--969.

\bibitem[{Freund and Schapire(1997)}]{freund1997decision}
Yoav Freund and Robert~E Schapire. 1997.
\newblock A decision-theoretic generalization of on-line learning and an
  application to boosting.
\newblock \emph{Journal of computer and system sciences}, 55(1):119--139.

\bibitem[{Hao et~al.(2019)Hao, Chen, Yu, Sun, and Wang}]{hao2019joie}
Junheng Hao, Muhao Chen, Wenchao Yu, Yizhou Sun, and Wei Wang. 2019.
\newblock Universal representation learning of knowledge bases by jointly
  embedding instances and ontological concepts.
\newblock In \emph{Proceedings of the ACM SIGKDD International Conference on
  Knowledge Discovery and Data Mining (KDD)}, pages 1709--1719. {ACM}.

\bibitem[{Hao et~al.(2020)Hao, Ju, Chen, Sun, Zaniolo, and
  Wang}]{hao2020biojoie}
Junheng Hao, Chelsea Ju, Muhao Chen, Yizhou Sun, Carlo Zaniolo, and Wei Wang.
  2020.
\newblock Bio-joie: Joint representation learning of biological knowledge
  bases.
\newblock In \emph{Proceedings of the 11st ACM Conference on Bioinformics,
  Computational Biology and Biomedicine (BCB)}. ACM.

\bibitem[{Ji et~al.(2015)Ji, He, Xu, Liu, and Zhao}]{ji2015knowledge}
Guoliang Ji, Shizhu He, Liheng Xu, Kang Liu, and Jun Zhao. 2015.
\newblock Knowledge graph embedding via dynamic mapping matrix.
\newblock In \emph{Proceedings of the Annual Meeting of Associations for
  Computational Linguistics (ACL)}, pages 687--696. The Association for
  Computer Linguistics.

\bibitem[{Ji et~al.(2020)Ji, Pan, Cambria, Marttinen, and Yu}]{ji2020survey}
Shaoxiong Ji, Shirui Pan, Erik Cambria, Pekka Marttinen, and Philip~S Yu. 2020.
\newblock A survey on knowledge graphs: Representation, acquisition and
  applications.
\newblock \emph{arXiv preprint arXiv:2002.00388}.

\bibitem[{Kingma and Ba(2014)}]{kingma2014adam}
Diederik~P Kingma and Jimmy Ba. 2014.
\newblock Adam: A method for stochastic optimization.
\newblock In \emph{International Conference on Learning Representations
  (ICLR)}.

\bibitem[{Koncel-Kedziorski et~al.(2019)Koncel-Kedziorski, Bekal, Luan, Lapata,
  and Hajishirzi}]{koncel2019text}
Rik Koncel-Kedziorski, Dhanush Bekal, Yi~Luan, Mirella Lapata, and Hannaneh
  Hajishirzi. 2019.
\newblock Text generation from knowledge graphs with graph transformers.
\newblock In \emph{Proceedings of the 2019 Conference of the North American
  Chapter of the Association for Computational Linguistics: Human Language
  Technologies, Volume 1 (Long and Short Papers) (NAACL)}, pages 2284--2293.
  Association for Computational Linguistics.

\bibitem[{Lehmann et~al.(2015)Lehmann, Isele, Jakob, Jentzsch, Kontokostas,
  Mendes, Hellmann, Morsey, Van~Kleef, Auer et~al.}]{lehmann2015dbpedia}
Jens Lehmann, Robert Isele, Max Jakob, Anja Jentzsch, Dimitris Kontokostas,
  Pablo~N Mendes, Sebastian Hellmann, Mohamed Morsey, Patrick Van~Kleef,
  S{\"o}ren Auer, et~al. 2015.
\newblock Dbpedia--a large-scale, multilingual knowledge base extracted from
  wikipedia.
\newblock \emph{Semantic Web}, 6(2):167--195.

\bibitem[{Li et~al.(2019)Li, Cao, Hou, Shi, Li, and Chua}]{li2019semi}
Chengjiang Li, Yixin Cao, Lei Hou, Jiaxin Shi, Juanzi Li, and Tat-Seng Chua.
  2019.
\newblock Semi-supervised entity alignment via joint knowledge embedding model
  and cross-graph model.
\newblock In \emph{Proceedings of the 2019 Conference on Empirical Methods in
  Natural Language Processing and the 9th International Joint Conference on
  Natural Language Processing (EMNLP-IJCNLP)}, pages 2723--2732. Association
  for Computational Linguistics.

\bibitem[{Nickel et~al.(2016)Nickel, Rosasco, Poggio
  et~al.}]{nickel2016holographic}
Maximilian Nickel, Lorenzo Rosasco, Tomaso~A Poggio, et~al. 2016.
\newblock Holographic embeddings of knowledge graphs.
\newblock In \emph{Proceedings of AAAI Conference on Artificial Intelligence
  (AAAI)}, pages 1955--1961. {AAAI} Press.

\bibitem[{Nickel et~al.(2011)Nickel, Tresp, and Kriegel}]{nickel2011three}
Maximilian Nickel, Volker Tresp, and Hans-Peter Kriegel. 2011.
\newblock A three-way model for collective learning on multi-relational data.
\newblock In \emph{Proceedings of the International Conference on Machine
  Learning (ICML)}, pages 809--816. Omnipress.

\bibitem[{Pei et~al.(2019)Pei, Yu, Hoehndorf et~al.}]{pei2019deg}
Shichao Pei, Lu~Yu, Robert Hoehndorf, et~al. 2019.
\newblock semi-supervised entity alignment via knowledge graph embedding with
  awareness of degree difference.
\newblock In \emph{Proceedings of the Web Confererence (WWW)}, pages
  3130--3136. {ACM}.

\bibitem[{Rebele et~al.(2016)Rebele, Suchanek, Hoffart, Biega, Kuzey, and
  Weikum}]{rebele2016yago}
Thomas Rebele, Fabian Suchanek, Johannes Hoffart, Joanna Biega, Erdal Kuzey,
  and Gerhard Weikum. 2016.
\newblock Yago: A multilingual knowledge base from wikipedia, wordnet, and
  geonames.
\newblock In \emph{Proceedings of the International Semantic Web Conference
  (ISWC)}, volume 9982 of \emph{Lecture Notes in Computer Science}, pages
  177--185. Springer.

\bibitem[{Shen et~al.(2017)Shen, Wu, Lei, Shang, Ren, and
  Han}]{shen2017setexpan}
Jiaming Shen, Zeqiu Wu, Dongming Lei, Jingbo Shang, Xiang Ren, and Jiawei Han.
  2017.
\newblock Setexpan: Corpus-based set expansion via context feature selection
  and rank ensemble.
\newblock In \emph{Joint European Conference on Machine Learning and Knowledge
  Discovery in Databases}, volume 10534 of \emph{Lecture Notes in Computer
  Science}, pages 288--304. Springer.

\bibitem[{Suchanek et~al.(2011)Suchanek, Abiteboul et~al.}]{suchanek2011paris}
Fabian~M Suchanek, Serge Abiteboul, et~al. 2011.
\newblock Paris: Probabilistic alignment of relations, instances, and schema.
\newblock \emph{Proceedings of the VLDB Endowment (PVLDB)}, 5(3).

\bibitem[{Sun et~al.(2017)Sun, Hu, and Li}]{sun2017cross}
Zequn Sun, Wei Hu, and Chengkai Li. 2017.
\newblock Cross-lingual entity alignment via joint attribute-preserving
  embedding.
\newblock In \emph{Proceedings of the International Semantic Web Conference
  (ISWC)}, pages 628--644. Springer.

\bibitem[{Sun et~al.(2018)Sun, Hu, Zhang, and Qu}]{sun2018bootstrapping}
Zequn Sun, Wei Hu, Qingheng Zhang, and Yuzhong Qu. 2018.
\newblock Bootstrapping entity alignment with knowledge graph embedding.
\newblock In \emph{Proceedings of the International Joint Conference on
  Artificial Intelligence (IJCAI)}, pages 4396--4402. International Joint
  Conferences on Artificial Intelligence Organization.

\bibitem[{Sun et~al.(2020{\natexlab{a}})Sun, Wang, Hu, Chen, Dai, Zhang, and
  Qu}]{sun2020alinet}
Zequn Sun, Chengming Wang, Wei Hu, Muhao Chen, Jian Dai, Wei Zhang, and Yuzhong
  Qu. 2020{\natexlab{a}}.
\newblock Knowledge graph alignment network with gated multi-hop neighborhood
  aggregation.
\newblock In \emph{Proceedings of AAAI Conference on Artificial Intelligence
  (AAAI)}, pages 222--229. {AAAI} Press.

\bibitem[{Sun et~al.(2019{\natexlab{a}})Sun, Wang, Hu, Chen, and
  Qu}]{sun2019transedge}
Zequn Sun, Jiacheng~Huang Wang, Wei Hu, Muhao Chen, and Yuzhong Qu.
  2019{\natexlab{a}}.
\newblock Transedge: Translating relation-contextualized embeddings for
  knowledge graphs.
\newblock In \emph{Proceedings of the International Semantic Web Conference
  (ISWC)}, pages 612--629. Springer.

\bibitem[{Sun et~al.(2020{\natexlab{b}})Sun, Zhang, Hu, Wang, Chen, Akrami, and
  Li}]{sun2020benchmark}
Zequn Sun, Qingheng Zhang, Wei Hu, Chengming Wang, Muhao Chen, Farahnaz Akrami,
  and Chengkai Li. 2020{\natexlab{b}}.
\newblock A benchmarking study of embedding-based entity alignment for
  knowledge graphs.
\newblock \emph{Proceedings of the VLDB Endowment}, 13:2326--2340.

\bibitem[{Sun et~al.(2019{\natexlab{b}})Sun, Deng, Nie, and
  Tang}]{sun2018rotate}
Zhiqing Sun, Zhi-Hong Deng, Jian-Yun Nie, and Jian Tang. 2019{\natexlab{b}}.
\newblock Rotate: Knowledge graph embedding by relational rotation in complex
  space.
\newblock In \emph{International Conference on Learning Representations
  (ICLR)}.

\bibitem[{Trouillon et~al.(2016)Trouillon, Welbl, Riedel, Gaussier, and
  Bouchard}]{trouillon2016complex}
Th\'eo Trouillon, Johannes Welbl, Sebastian Riedel, \'Eric Gaussier, and
  Guillaume Bouchard. 2016.
\newblock {Complex embeddings for simple link prediction}.
\newblock In \emph{Proceedings of the International Conference on Machine
  Learning (ICML)}, volume~48, pages 2071--2080. PMLR.

\bibitem[{Trsedya et~al.(2019)Trsedya, Qi, and Zhang}]{distiawanTrsedya2019}
Bayu~Distiawan Trsedya, Jianzhong Qi, and Rui Zhang. 2019.
\newblock Entity alignment between knowledge graphs using attribute embeddings.
\newblock In \emph{Proceedings of the AAAI Conference on Artificial
  Intelligence (AAAI)}, pages 297--304. {AAAI} Press.

\bibitem[{Vrande{\v{c}}i{\'c} and Kr{\"o}tzsch(2014)}]{vrandevcic2014wikidata}
Denny Vrande{\v{c}}i{\'c} and Markus Kr{\"o}tzsch. 2014.
\newblock Wikidata: a free collaborative knowledge base.
\newblock \emph{Communications of ACM}, 57(10):78–85.

\bibitem[{Wang et~al.(2017)Wang, Mao, Wang, and Guo}]{wang2017knowledge}
Quan Wang, Zhendong Mao, Bin Wang, and Li~Guo. 2017.
\newblock Knowledge graph embedding: A survey of approaches and applications.
\newblock \emph{IEEE Transactions on Knowledge and Data Engineering},
  29(12):2724--2743.

\bibitem[{Wang et~al.(2014)Wang, Zhang, Feng, and Chen}]{wang2014knowledge}
Zhen Wang, Jianwen Zhang, Jianlin Feng, and Zheng Chen. 2014.
\newblock Knowledge graph embedding by translating on hyperplanes.
\newblock In \emph{Proceedings of AAAI Conference on Artificial Intelligence
  (AAAI)}, pages 1112--1119. {AAAI} Press.

\bibitem[{Wang et~al.(2018)Wang, Lv, Lan, and Zhang}]{wang2018cross}
Zhichun Wang, Qingsong Lv, Xiaohan Lan, and Yu~Zhang. 2018.
\newblock Cross-lingual knowledge graph alignment via graph convolutional
  networks.
\newblock In \emph{Proceedings of the Conference on Empirical Methods in
  Natural Language Processing (EMNLP)}, pages 349--357. Association for
  Computational Linguistics.

\bibitem[{Wolpert(1992)}]{wolpert1992stacked}
David~H Wolpert. 1992.
\newblock Stacked generalization.
\newblock \emph{Neural networks}, 5(2):241--259.

\bibitem[{Yang et~al.(2015)Yang, Yih, He, Gao, and Deng}]{yang2014embedding}
Bishan Yang, Wen-tau Yih, Xiaodong He, Jianfeng Gao, and Li~Deng. 2015.
\newblock Embedding entities and relations for learning and inference in
  knowledge bases.
\newblock \emph{International Conference on Learning Representations (ICLR)}.

\bibitem[{Yang et~al.(2019)Yang, Zou, Shi, Lu, Lin, and Sun}]{yang2019aligning}
Hsiu-Wei Yang, Yanyan Zou, Peng Shi, Wei Lu, Jimmy Lin, and Xu~Sun. 2019.
\newblock Aligning cross-lingual entities with multi-aspect information.
\newblock In \emph{Proceedings of the 2019 Conference on Empirical Methods in
  Natural Language Processing and the 9th International Joint Conference on
  Natural Language Processing (EMNLP-IJCNLP)}, pages 4430--4440. Association
  for Computational Linguistics.

\bibitem[{Zhang et~al.(2019)Zhang, Sun, Hu, Chen, Guo, and Qu}]{zhang2019multi}
Qingheng Zhang, Zequn Sun, Wei Hu, Muhao Chen, Lingbing Guo, and Yuzhong Qu.
  2019.
\newblock Multi-view knowledge graph embedding for entity alignment.
\newblock In \emph{Proceedings of the International Joint Conference on
  Artificial Intelligence (IJCAI)}, pages 5429--5435. International Joint
  Conferences on Artificial Intelligence Organization.

\bibitem[{Zhang et~al.(2020)Zhang, Chen, and Bui}]{zhang2020diagnostic}
Tianran Zhang, Muhao Chen, and Alex Bui. 2020.
\newblock Diagnostic prediction with sequence-of-sets representation learning
  for clinical event.
\newblock In \emph{Proceedings of the 18th International Conference on
  Artificial Intelligence in Medicine (AIME)}. Springer.

\end{thebibliography}
